# Representation of 2D frameless visual space as a neural manifold and its information geometric interpretation

Debasis Mazumdar

Centre for Development of Advanced Computing

Block-GP,Plot-E-2/1,Sector-V

Saltlake, Kolkata-700091.India.

e-mail debases.mazumdar@cdac.in

Abstract

Representation of 2D frameless visual space as neural manifold and its modelling in the frame work of information geometry is presented. Origin of hyperbolic nature of the visual space is investigated using evidences from neuroscience. Based on the results we propose that the processing of spatial information, particularly estimation of distance, perceiving geometrical curves etc. in the human brain can be modelled in a parametric probability space endowed with Fisher-Rao metric. Compactness, convexity and differentiability of the space is analysed and found that they obey the axioms of G space, proposed by Busemann. Further it is shown that it can be considered as a homogeneous Riemannian space of constant negative curvature. It is therefore ensured that the space yields geodesics into it. Computer simulation of geodesics representing a number of visual phenomena and advocating the hyperbolic structure of visual space is carried out. Comparison of the simulated results with the published experimental data is presented.

1.Introduction

Visual sensory perception differs from other sense modalities in the sense that we see the physical scene as a boundless three dimensional extent articulated into objects having size, form and positions. Visual space on the other hand is the product of a series of coherent, self-organized and complex processes evoked into retina and followed by different cellular layers in the visual path and ultimately formed at the primary visual cortex. Digressing from the notion of subjective nature of visual space, Helmholtz pointed out that visual space is geometrical in nature[1]. Of course there are personalized psychological and neurological parameters of the subject having strong influence on the geometric structure of the visual space. Luneburg was the first to propose a complete mathematical model of the binocular visual space perception[2],[3].To explain several visual phenomena, namely the Helmholtz's horopter [1], Ame's distorted room[4],parallel and distance alleys reported by Hillebrand and Blumenfeld [5],[6],Luneburg developed his model assuming, (a) the visual space is frameless, i.e. no framework like wall or ground is visible,(b) all percepts are localized with respect to the self (egocentric). To explain the experimental data, particularly the parallel and distance alleys, it is further hypothesized that, (c) it must be a homogeneous Riemannian metric space of constant negative curvature (hyperbolic) in which free mobility of the objects are ensured. Luneburg's mathematical analysis of the binocular visual space was a great achievement in its time and impacted the research community. The theory was furthered by Blank, Foley, Nishikawa, Indow [5], [7], [ 8], [9] and many others. Also the experimental support to the theory has been reported [10].Local transformations between the physical and visual space was formulated by [11], [12]. All these studies are based on the metric properties of the visual space and are within the Luneburg's model of hyperbolic visual space. Remarkably, till today there is lack of theoretical studies to explain the neurological origin of the hyperbolic nature of the visual space. Neither any mathematical description of the neural coding of the visual space is available. Present paper models the neuro-manifold where an input $\xi$ is assumed to be coded in the mean firing rate $f(\xi)$, known as the tuning curve. Gaussian functions are chosen to represent the tuning curve and collecting experimental data from neuroscience it is further considered that the spatial information of the physical space is encoded in the two parameters of the tuning curve, namely its peak $\mu_\xi$ and the scale factor $\sigma_\xi$ . The visual space is then modelled as a two dimensional parametric space of Gaussian random variables, parameterized by ($\mu_\xi, \sigma_\xi$) and the psychological distance function is considered to be the Fisher-Rao distance metric.. Mathematically the space is a homogeneous Riemannian space (locally Euclidean) with constant negative curvature, allowing free mobility of rigid objects. The compactness, convexity and differentiability of the space is analysed to explain that geodesics can be calculated in the space according to the postulates of Busemann[13].It is further demonstrated that, under Mobius transformation the space is homeomorphic and conformal too to the space considered by Luneburg using Poincare disk model as the Euclidean map of the visual space. Equations of geodesics in the proposed information geometric are obtained. Computer simulations of a number of well-known visual phenomena (Helmholtz's horopters, parallel and distance alleys) are performed. Comparison of simulated results with the published experimental data is presented.

2. Coordinate systems for physical and visual space of binocular vision.

In order to give a geometric characterization of a stimulus configuration, suitable coordinate system should be defined to represent physical and visual space. Following Luneburg [3] we choose the bipolar coordinate system $Q$ $\lambda, \psi$ to represent the stimulus in the physical space Fig.1a. $\lambda$ is the bipolar parallax approximating the angle of convergence of the visual axes. $\psi$ is the bipolar latitude which approximates the average inclinations of the two visual axes with respect to the median plane. The axes of reference are the x and y axis. The y-axis passing through the centre of rotation of the eyes and oriented positively to the right. The x-axis is orthogonal to the y-axis at the midpoint between the centres of rotation of the two eyes. To represent the visual space a local Cartesian coordinate system is defined as $\rho, \phi$ and with the axes of reference $\mu$ and $\sigma$ Fig.1b. The coordinate system is a Euclidean map and conformal to the coordinate system of the physical space. The neurological interpretation of $\mu$ and $\sigma$ will be described in the next section. The Cartesian coordinates $\mu, \sigma$ are related to the polar coordinates $\rho, \phi$ as,

$$\mu = \rho \cos(\phi) \qquad (1.a)$$

$$\sigma = \rho \sin(\phi) \qquad (1.b)$$

It is assumed that the axes x and y in the physical space coincides directly to the local coordinate axes $\mu$ and $\sigma$ in the visual space. The bipolar latitude $\psi$ of the physical space equals the polar angle $\phi$ of the visual space.

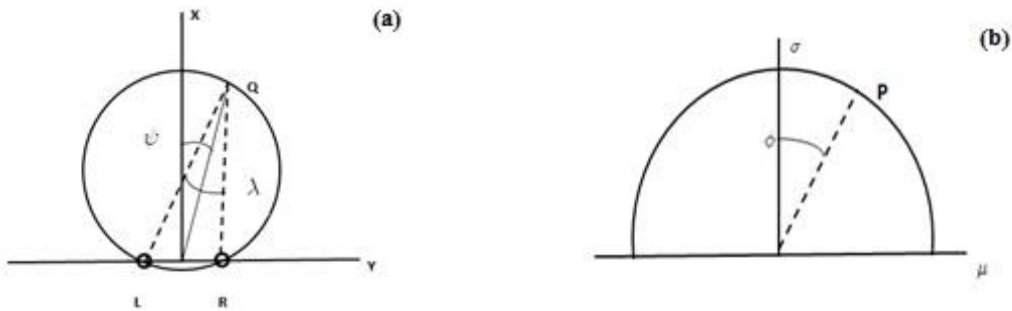

Fig.1 Coordinates of the physical and visual space. (a) Represents the bipolar coordinate system describing the physical space.(b) Represents the local Cartesian coordinate and the corresponding polar coordinate system describing the visual space. All the symbols are described in the text.

The stimulus in the horizontal plane of the visual space are either represented by the polar coordinates ($\rho, \phi$) or the Cartesian coordinates ($\mu, \sigma$).The origin of the coordinate system is assumed to be the apparent centre of observation. It is further assumed that the physical axes (x, y) coincides with the axes ($\mu, \sigma$) in both the spaces. Stimulus point subtending the bipolar parallax $\lambda$ = constant in the physical space are perceived at a distance $\rho$ = constant in the visual space. Therefore the relation between the physical and the visual space boils down to the relation between the bipolar parallax $\lambda$ and the radial distance $\rho$. The relation between the variables in the physical and visual space is defined by Luneburg by a simple mapping function [2] which is utilized for the computer simulations of different visual phenomena and presented in the subsequent sections.

## 2A. Representation of metric of the visual space encoded as neural assemblies in the brain.

To compute the psychometric distance of the visual space we consider a one dimensional continuous neural field model [14],[15],[16],with uniform density which is very high ($10^{11}$ neurons and $10^{14}$ connections). Let the ensemble of neurons are coding a variable $\xi$, representing the position of the stimulus in the visual space. The activity of neurons is represented by the tuning curve,

$$f(\xi) = r_{max} \exp\left(-\frac{1}{2}\left(\frac{\xi - \mu_\xi}{\sigma_\xi}\right)^2\right) \qquad . \qquad (2)$$

The tuning curve represents the average firing rate as a function of the stimulus position $\xi$ within a region of the visual field where the neurons responded to the stimulus. The region is technically called the receptive field (RF) of the neuron and is usually represented by difference of Gaussian function (DOG) having distinct excitatory centre and inhibitory surround region. The tuning curve has its peak at $\mu_\xi$, the preferred position of the stimulus evoking maximum average firing rate $r_{max}$. The scale factor $\sigma_\xi$ represents the inter trial variations of the neural responses even when same stimulus is presented before the subject in number of trials. The variability in the neural responses includes variable levels of arousal, attentional variability of the subject and various biophysical processes which influence neuronal firing. To model the encoding process of representing object locations by the tuning curve we consult the results of several neuro scientific experiments. In variety of sensory system it is found that tuning curve provides the functional description of response of neuron towards a stimulus. It includes orientation columns in the vertebrate visual cortex. place cells in the hippocampus, wind-detecting neurons in the cricket cercal system [16]. To develop the suitable psychological metric function which is close to the neural processing of the task of determination of the distance between two point stimulus we first explain briefly the experiment conducted by Oleksiak et al. [17]. The experimental results show that the estimation of distance depends on the encoding condition and it causes variation in the functional shape of the tuning curve and the receptive field. In the experiment nine human observers with normal or corrected to normal vision took part. The subjects were instructed to estimate the horizontal separation between two dots by placing a movable cursor. The subjects gave their judgments of estimation of the gap in two distinct visual condition, namely the fixate and the saccade. In fixate trial the registration of the stimuli was done through peripheral vision while in saccade trial it was done through foveal vision. In each case estimation bias (error) was noted. Interestingly the measurement performance in fixate and saccade vision differed in three measures designated as, (i) absolute error,(ii) signed error and (iii) variability of responses. To analyse the experimental result it is concluded that in foveal and peripheral vision the neurons tuning curve width varies which directly translated into the size of the receptive field of the neurons. The peripheral vision encodes the positional information with a larger RF while in foveal vision the information is sampled through RF of relatively smaller size. The mutual interaction of the RF of the two dots generates error in estimation of the gap between two stimuli. More accurately, The peak of each of the RF is considered to represent the location of the respective dots. Superposition of two RFs modulate their shapes and the shift of their peak generates the estimation error. It is to be noted here that similar explanations were published earlier by Blakemore et al. [18]. Being motivated by these experimental results we consider that the locational information of objects is encoded by two features of neuronal firing, namely the maximum firing rate and the maximum change of firing rate. In functional form the two features are represented as the peak of the Gaussian tuning curve ($\mu_\xi$) and its scale factor ($\sigma_\xi$) respectively. It is to be noted here that mean of the tuning curve represents the position of the preferred stimulus. The standard deviation is significant as well, because it determines the highest slope of the tuning curve to encode the maximum change in the firing rate. For two infinitesimally close stimuli the slope of the tuning curve is used to discriminate them. Therefore the proposed information theoretic model of the visual space is a half plane with a frames of reference defined by mean $\mu$ and standard deviation $\sigma$ of the tuning curves,

$$H_F = \{(\mu,\sigma) \in R^2 \mid \sigma > 0\} . \qquad (3)$$

The half plane associates each point in the upper half plane of $R^2$ with a univariate Gaussian pdf p (r;$\Omega$). The univariate Gaussian pdf is actually the tuning curve representing the functional description of neuron's firing state. r is a stochastic variable representing the firing rate of the neuron and is supported on a region X representing the size of the RF. $\Omega = (\omega^1, \omega^2)$ is the parameter characterising the tuning curve. In the present model $\omega^1 = \mu$ and $\omega^2 = \sigma$. The tuning curves are actually represented by probability distribution function which obeys the normalization condition,

$$\int_X p(r,\Omega)dr = 1 . \qquad (4)$$

Expectation value of any observable in this space can be represented as,

$$\langle O \rangle = \int_X p(r,\Omega)O(r,\Omega)dr \qquad (5)$$

The distance between two points $p_1 = (\mu_1,\sigma_1)$ and $p_2 = (\mu_2,\sigma_2)$ in the upper half plane H measures the dissimilarity between the Gaussian functions associated with the respective tuning curve. Therefore the dissimilarity between two infinitesimally separated stimuli represented by their tuning curves $p(r,\Omega)$ and $p(r,\Omega + d\Omega)$ can be calculated using the Kullback- Leibler divergence,

$$d_{kl} = \int_r p(r,\Omega) \ln p(r,\Omega) dr - \int_r p(r,\Omega) \ln p(r,\Omega + d\Omega) dr . \qquad (6)$$

Expanding the integrand of the second integral into Taylor series up to the second order terms we get,

$$d_{kl} \cong -\frac{1}{2} \frac{\partial}{\partial \Omega^\mu} \int_r [p(r,\Omega)] d\Omega^\mu + \frac{1}{2} g_{\mu\nu} d\Omega^\mu d\Omega^\nu . \qquad (7)$$

$$\text{Where, } g_{\mu\nu}(\phi) = \int_r p(r,\Omega) \frac{\partial \gamma}{\partial \Omega^\mu} \frac{\partial \gamma}{\partial \Omega^\nu} dr = \langle \partial_\mu \gamma \partial_\nu \gamma \rangle . \qquad (8)$$

$\gamma(r,\Omega) = -\ln p(r;\Omega)$ is termed as the spectrum. $g_{\mu\nu}$ is the metric tensor, technically known as Fisher-Rao information matric [19] and it defines the inner product in the upper half plane H as,

$$\langle l_1, l_2 \rangle = l_1^T (g_{\mu\nu}) l_2 \text{ with } || l_1 || = \sqrt{\langle l_1, l_1 \rangle} \text{ and } || l_2 || = \sqrt{\langle l_2, l_2 \rangle} .$$

The differential distance is defined as,

$$ds_F^2 = \sum_{\mu,\nu=1}^{2} g_{\mu\nu} d\Omega^\mu d\Omega^\nu . \qquad (9)$$

Various components of the metric tensor $g_{\mu\nu}$ when computed [20], are found as $g_{\mu\mu} = \frac{1}{\sigma^2}, g_{\sigma\sigma} = \frac{2}{\sigma^2}$ and $g_{\mu\sigma} = g_{\sigma\mu} = 0$. Therefore the expression of the differential distance element of the visual space is,

$$ds_F^2 = \frac{d\mu^2 + 2d\sigma^2}{\sigma^2} . \qquad (10)$$

The Fisher-Rao distance between the two points ( $\mu_1, \sigma_1$ ) and ( $\mu_2, \sigma_2$ ) is given by [20],

$$d_F = \sqrt{2} \ln \left( \frac{\kappa((\mu_1,\sigma_1),(\mu_2,\sigma_2)) + (\mu_1^2 - \mu_2^2) + 2(\sigma_1^2 + \sigma_2^2)}{4\sigma_1\sigma_2} \right). \qquad (11)$$

Where, $\kappa((\mu_1,\sigma_1),(\mu_2,\sigma_2)) = \left[ ((\mu_1 - \mu_2)^2 + 2(\sigma_1 - \sigma_2)^2)((\mu_1 - \mu_2)^2 + 2(\sigma_1 + \sigma_2)^2) \right]^{\frac{1}{2}} .$

Using Eq.11 we computed the absolute error of distance estimation as the difference between the geodesic distance ($d_F$) between the two tuning curves and the true horizontal separation of the two dots ($d_u$) expressed as,

$$\varepsilon = |d_F - d_u|. \qquad (12)$$

Following [17], the horizontal separation between the two dots is varied ranging 2 to 16 deg at an interval of 2 deg of visual angle . The position of the more foveal dot ($\mu_1$) is assigned to zero while the location of the more peripheral of the two dots ($\mu_2$) is set to the value of their separation. The two free parameters $\sigma_1$ and $\sigma_2$ are computed in the following way. We first postulate that the subject assigns increasing attention towards the foveal dot when the separation with the peripheral dot increases. Alteration of tuning curve with different degrees of attention was studied earlier [21],[22],[23] and it was reported that in area V4 the neuronal tuning curves attain gradually decreasing width with increasing degree of attention towards a particular stimulus. Therefore it is expected that the scale factor of the tuning curve corresponding to the more foveal dot will be a decreasing function of the horizontal separation of the two dots. We first determine the value of the scale factor $\sigma_2$ corresponding to the more peripheral dot using a linear relation between the size of the tuning curve and the eccentricity of the stimulus, which is very close to the rough estimation of the RF size reported in [24 ],[25].

$$\sigma_2 = \frac{E^{\circ}_{cc} + \alpha}{\beta}. \qquad (13)$$

The value of $\alpha = 126.540$ and $\beta = 167.785$ resulted the closest fit with the data reported in [17] which corresponds to the saccade trial. In the saccade trial the ratio of separation by eccentricity is found to be one. Naturally the separation of dots corresponds to the eccentricity. Variation of $\sigma_2$ with the separation of the dots are presented in Fig.2b.The dotted line corresponds to the values computed using Eq.13 and used in the simulation for obtaining the closest match with the data presented in [17] and the solid curve represents the best fit line. After determining the value of $\sigma_2$, we adjusted the value of $\sigma_1$ manually to get the closest value of the data reported in [17]. Variation of $\sigma_1$ as a function of $d_u$ is presented in Fig.2a. The absolute error computed as a function of eccentricity is represented in Fig.2c and shows close resemblance to the experimental result. Based on this result we conclude that the Fisher information metric is indeed the appropriate metric of the visual space.

## 2B. Intrinsic topological properties of the frameless visual space in the information geometric framework.

In order to find the topological properties of the visual space, we first compute the components of Christoffel symbol $\Gamma^{x}_{yz}$ as they carry significant information regarding the geometrical properties of any Riemannian space. For any n dimensional Riemannian manifold there are N=$\frac{n^2(n+1)}{2}$ numbers of independent components of Christoffel symbols which can be computed using the equation,

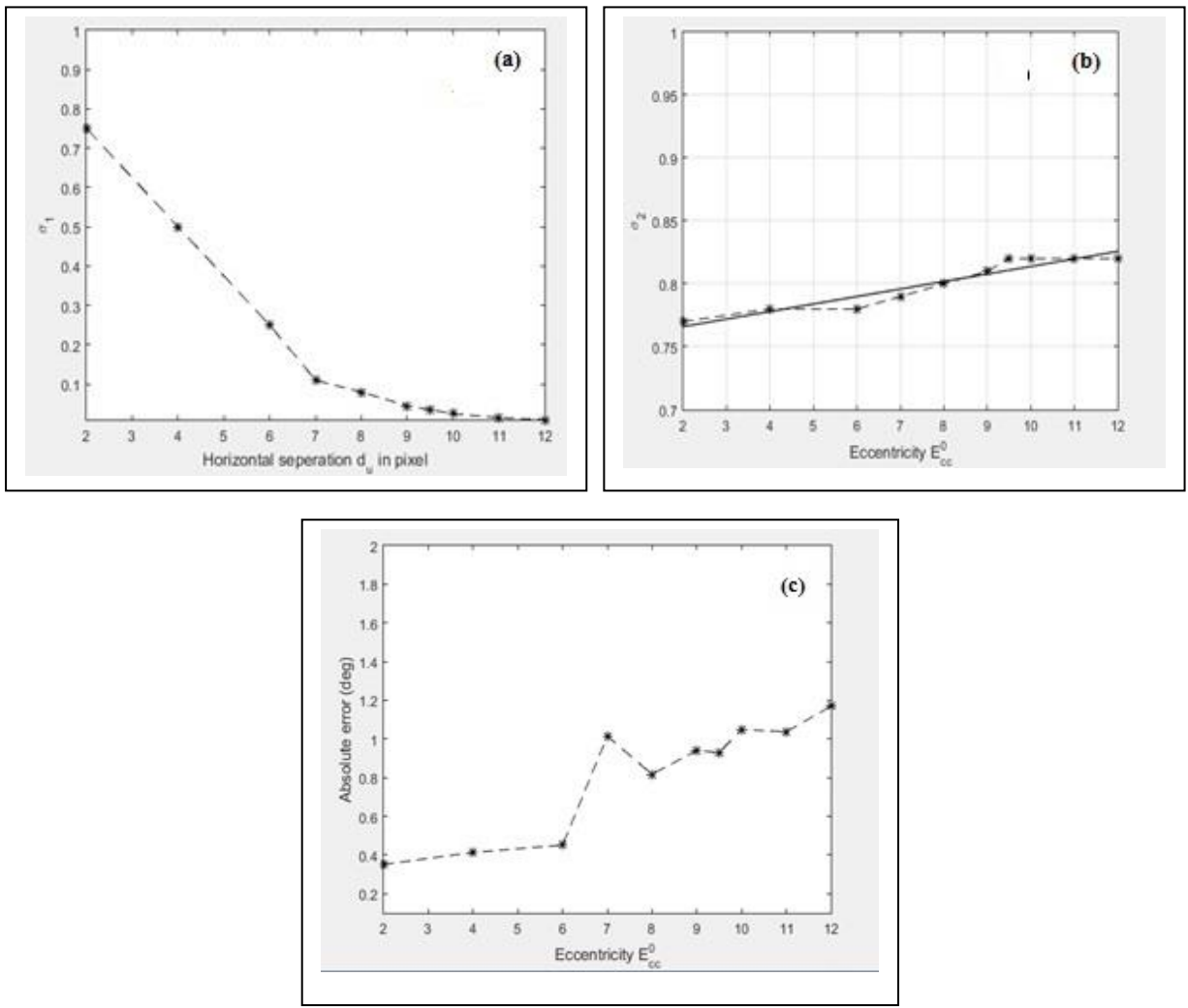

Fig.2 Simulation of perceptual errors in perceiving horizontal distances between two point stimuli.(a) Variation of the scale factor ($\sigma_1$) of the tuning curve evoked by the more foveal dot with horizontal separation of the stimuli.(b) Variation of the scale factor of the tuning curve evoked by the distant dot. The dotted curve represents the computed values of $\sigma_2$ and the solid line represent the linear fit of the data.(c) Variation of absolute error with the eccentricity of the dots.

$$\Gamma^{\mu}_{\lambda\sigma} = \frac{1}{2} g^{\mu\rho} \left[ \frac{\partial g_{\lambda\rho}}{\partial x^{\sigma}} + \frac{\partial g_{\rho\sigma}}{\partial x^{\lambda}} - \frac{\partial g_{\sigma\lambda}}{\partial x^{\rho}} \right]. \tag{14}$$

The two dimensional Fisher information space will have six independent components of Christoffel symbols and their computed values using Eq.14 gives, $\Gamma^{\mu}_{\mu\mu} = \Gamma^{\mu}_{\sigma\sigma} = \Gamma^{\sigma}_{\mu\sigma} = 0, \Gamma^{\mu}_{\mu\sigma} = \Gamma^{\sigma}_{\sigma\sigma} = -\frac{1}{\sigma}$ and $\Gamma^{\sigma}_{\mu\mu} = \frac{1}{2\sigma}$. The scalar curvature of any Riemannian space is defined as,

$$R = g^{ij} R_{ij}. \tag{15}$$

Where $R_{ij}$ is the component of Ricci curvature tensor. Expanding the Eq.15 using the values of the component of metric tensor $g^{ij}$ and using Einstein's summation convention we can write the scalar curvature tensor for the two dimensional Fisher information space as,

$$R = \sigma^2 R^{\sigma}_{\mu\sigma\mu} + \frac{\sigma^2}{2} R^{\mu}_{\sigma\mu\sigma} \tag{16}$$

The Riemann curvature tensors are obtained as, $R^{\mu}_{\sigma\mu\sigma} = -\frac{1}{\sigma^2}$ and finally the scalar curvature $R = \sigma^2 \left( -\frac{1}{2\sigma^2} \right) + \frac{\sigma^2}{2} \left( -\frac{1}{\sigma^2} \right) = -1$. Therefore the visual space endowed with the Fisher-Rao information metric is a Riemannian space of constant curvature -1 or it is a homogeneous hyperbolic space. It is worth noting here that, in any homogeneous Riemannian space of constant curvature the form and localization are completely uncorrelated and ensures the free mobility of rigid bodies as per the Helmholtz-Lie proposition [26],[27]. The experimental verification of the curvature of the visual space is done by many researchers and there are wide variations in their reported results. Blank[12] performed an ingenious experiment using point light sources to form an isosceles triangle in the horizontal plane. In Euclidean space the line joining the mid points of two sides of a triangle is equal in length to half the third side while in hyperbolic space it is smaller and in spherical space it is larger. Out of seven observers, judgement of six observers confirms the negative sign of the curvature, i.e. hyperbolic nature of the space while one observer does not exhibit significant curvature. Hagino and Yoshioka [28] arranged light points $Q_i$ along a horizontal axis and instructed five subjects to arrange another set of light points to form circles centered around $Q_i$s. Analysing the result according to Luneburg model they reported that the measurement of curvature based on the data is found negative in all cases while the value of the curvature is found varying depending on $Q_i$s and the radius of the circles. These two experiments are conducted in closed environment and can be regarded as representing frameless visual space in dark room condition. More recently Koenderink et al.[29] reported their result of measurement of intrinsic curvature of the visual space based on the Gauss's original definition, i.e. the angular excess in a triangle equals the integral curvature over the area of the triangle. The experiment was conducted in the natural environment, an open field under bright day light condition with everything in plain sight. The result shows that the curvature varies from elliptic in near space to hyperbolic in far space and parabolic at very large distance. Although the experimental results are not decisive and shows strong dependencies on the experimental conditions and the nature of the stimuli it can be said assertively that the frameless visual space evoked in the dark room condition is endowed with constant negative curvature. Though we don't have adequate neuroscientific data but logically it can be concluded that in the open field condition presence of different background visual cues distorts the tuning curves. Presence of many noisy tuning curves perturb the visual space and make it deviated from being a hyperbolic space of constant negative curvature. Another important topological property is the compactness of the visual space. The sequential compactness of the space of Gaussian random variables, representing the tuning curves cannot be proved almost surely. For example the sequential compactness will fall if we chose the parameter space as an open interval (0,1) on real line. Since the boundary points are left out. However we can realize a weaker convergence as described. Let $\{x_n\}$ be a bounded random sequence in the visual space. In general it is not true that there exist a subsequence of it that converges to a constant. But a weaker convergence can be chosen following Prohorov's theorem. It can be shown that a uniformly bounded sequence of random variables has a subsequence convergence in distribution. Within this constraint the visual space can be considered as (weakly) sequentially compact. Topologically any bounded infinite set of sequences in the visual space has an accumulation point. Extending the discussion we can further conclude that the hyperbolic lines, hyperbolic rays and hyperbolic line segments are convex in the Euclidean map like upper half plane or the Poincare disc [32].Although the convexity of the union of close intervals is not ensured. As an example we can consider the parameter values 1 and 2 then in the parameter space [0,1] ∪ [2,3] fails to attain internal convexity. The space is finally differentiable too. According to Busemann any space possessing all these properties yields geodesics [13],[26].The next important issue is the homeomorphism between different Euclidean maps of the hyperbolic visual space. The Euclidean map represented by the upper half plane model based on the Fisher-Rao metric $ds_F^2 = \frac{d\mu^2 + 2d\sigma^2}{\sigma^2} = \frac{|d\omega|^2}{\sigma^2}$ (where $\omega = \mu + i\sqrt{2}\sigma$) is homeomorphic with the metric of the Poincare disk model, $\frac{4(d\alpha^2 + d\beta^2)}{[1-(\alpha^2+\beta^2)]^2} = \frac{4|dz|^2}{[1-|z|^2]^2}$ ,where $z = \alpha + i\beta$. The latter is used in the Luneburg's model to describe different visual curves. It is easy and straightforward to verify that under the Mobius transformation $\omega = i\frac{(1+z)}{(1-z)}$ the Fisher-Rao information metric is homeomorphic with the metric of Poincare disc model apart from a constant factor of 2. Naturally all the curves derived in one model are transformable to the other conformally using the aforementioned Mobius transformation. Therefore Fisher-Rao information geometric upper half space and the Poincare disk are the two Euclidian map of the visual space which are homeomorphic. In the following we shall first derive the geodesic equation in the upper half space and subsequently analyse them after transforming in the Poincare disk model as various properties of the geodesics can be presented comfortably in the latter model both mathematically as well as visually.

3.A. Interpreting Horopter as the Geodesics of the Information Geometric Space.

In visual science, research on interpreting appearance of fixated objects single and apparently straight curves has long history [30].It is attributed to Helmholtz [1] to systematically study the fact that curves generating the impression of straightness are not straight at all distances from the observer. They are concave toward the observer at near distances and convex at far. At some intermediate distance they appear straight in the vicinity of the median Fig.3. The curves are termed as horopters or frontoparallel lines. They are defined as the locus of the points in the physical space, projected as image onto corresponding points in the two retinae. Physiologically, corresponding points are the pairs of loci in the two retinae evoking a fused image in cyclopean vision. Geometrically they are the points at equiangular deviation from the foveae. Therefore horopters are the Vieth-Muller circles.

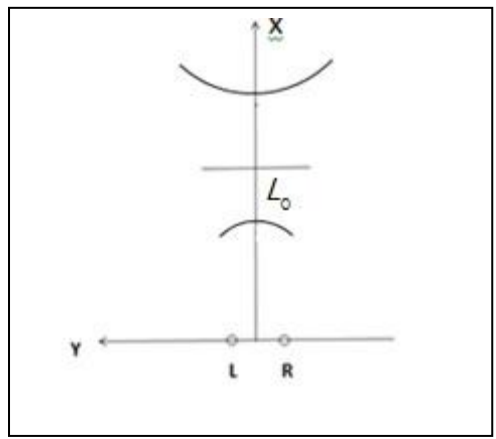

Fig.3. Frontoparallel curves at different distances from the observer. The X-axis is in the direction of the median line of the two eyes. The Y-axis is in the horizontal direction and passing through the optical centres of the left (L) and right ( R ) eyes. $L_o$ is the vertical distance of the straight line horopter from the origin.

Luneburg [2] interpreted them as the geodesics of the hyperbolic space which are symmetrical about the axis defining the median. The geodesic equation derived by Luneburg used the metric of the Poincare disk model as mentioned in the previous section. The interpretation of the horopters as geodesics of the information geometric space can be obtained by solving the geodesic equation,

$$\frac{d^2 X_k}{ds^2} + \sum_{i=1, j=1}^{2} \Gamma_{i,j}^{k} \frac{dx_i}{ds} \frac{dx_j}{ds} = 0 \ . \tag{17}$$

Using the Fisher-Rao metric with index k =1,2 and setting the variables $x_1 = \mu$ and $x_2 = \sigma$ we obtain,

$$\frac{d^2\mu}{ds^2} + \left[ \Gamma_{\mu\mu}^{\mu} \frac{d\mu}{ds}\frac{d\mu}{ds} + \Gamma_{\mu\sigma}^{\mu} \frac{d\mu}{ds}\frac{d\sigma}{ds} + \Gamma_{\sigma\mu}^{\mu} \frac{d\sigma}{ds}\frac{d\mu}{ds} + \Gamma_{\sigma\sigma}^{\mu} \frac{d\sigma}{ds}\frac{d\sigma}{ds} \right] = 0 \text{ for k=1}$$

and $$\frac{d^2\sigma}{ds^2} + \left[ \Gamma_{\mu\mu}^{\sigma} \frac{d\mu}{ds}\frac{d\mu}{ds} + \Gamma_{\mu\sigma}^{\sigma} \frac{d\mu}{ds}\frac{d\sigma}{ds} + \Gamma_{\sigma\mu}^{\sigma} \frac{d\sigma}{ds}\frac{d\mu}{ds} + \Gamma_{\sigma\sigma}^{\sigma} \frac{d\sigma}{ds}\frac{d\sigma}{ds} \right] = 0 \text{ for k=2.}$$

Finally replacing the different components of Christoffel symbols, derived earlier in Eq.14. we get the differential equations,

$$\frac{d^2\mu}{ds^2} - \frac{2}{\sigma} \frac{d\mu}{ds} \frac{d\sigma}{ds} = 0 \quad \text{for k=1,} \tag{18}$$

and $$\frac{d^2\sigma}{ds^2} + \frac{1}{\sigma}\left[ \frac{1}{2}\left(\frac{d\mu}{ds}\right)^2 - \left(\frac{d\sigma}{ds}\right)^2 \right] = 0 \ . \tag{19}$$

To solve the above coupled differential equations we consider the function,

$$P = 2\sigma \frac{d\sigma}{ds} + \left(\frac{d\mu}{ds}\right)^{-1} + \mu \text{ and obtain,}$$

$$\frac{dp}{ds} = \frac{2\left(\frac{d\sigma}{ds}\right)^2 \frac{d\mu}{ds} + 2\sigma\left(\frac{d^2\sigma}{ds^2}\right)\frac{d\mu}{ds} - 2\sigma\left(\frac{d^2\mu}{ds^2}\right)\frac{d\sigma}{ds} + \left(\frac{d\mu}{ds}\right)^3}{\left(\frac{d\mu}{ds}\right)^2} \tag{20}$$

Using Eq. 18 and Eq.19 respectively we further calculate,

$$2\sigma \frac{d^2\mu}{ds^2}\left(\frac{d\sigma}{ds}\right) = 4\left(\frac{d\sigma}{ds}\right)^2 \frac{d\mu}{ds} \quad , \tag{21}$$

and $$\left(\frac{d\mu}{ds}\right)^3 = -2\sigma \frac{d^2\sigma}{ds^2}\frac{d\mu}{ds} + 2\left(\frac{d\sigma}{ds}\right)^2 \frac{d\mu}{ds} \ . \tag{22}$$

When substitute them in Eq.20 we get the equation,

$$\frac{dp}{ds} = \frac{d}{ds}\left[2\sigma\left(\frac{d\sigma}{ds}\right)\left(\frac{d\mu}{ds}\right)^{-1} + \mu\right] = 0. \qquad (23)$$

Therefore $\left[2\sigma\left(\frac{d\sigma}{ds}\right)\left(\frac{d\mu}{ds}\right)^{-1} + \mu\right] = K$. Where $K$ is a constant. With simple calculation it reduces to,

$$\sigma^2 + \left[\frac{\mu}{2} - K_1\right]^2 = K_2, \qquad (24)$$

where $K_1 = \frac{K}{2}$ and $K_2$ is another constant. Eq.24 can be further simplified as,

$$2\sigma^2 + \mu^2 - C\mu = 1, \qquad (25)$$

In reaching Eq.25 from Eq.24 we substituted $C = 4K_1$.

Eq.25 represents the equation of a circle whose centre lies on the $\mu$ axis and intersects the axis perpendicularly. These are the geodesic of the visual space with Fisher-Rao metric. With simple algebraic calculation we can rewrite Eq.25 as,

$$\left(\mu - \frac{C}{2}\right)^2 + 2\sigma^2 = r^2. \qquad (26)$$

Where $r = \left(1 + \frac{C^2}{4}\right)^{\frac{1}{2}}$. We further introduce the variable $\omega = \mu + i\sqrt{2}\sigma$, and $\omega_0 = \frac{C}{2}$ to write Eq.26 as,

$$(\omega_1 - \omega_0)(\omega_1 - \omega_0)^* = r^2. \qquad (27)$$

Applying the Mobius transformation as defined above, Eq.27 transforms to,

$$\frac{K_3}{4}(\alpha^2 + \beta^2) - 1 = c\alpha. \qquad (28)$$

Eq.28 represents the locus of the Helmholtz horopter obtained by Luneburg [3] using Poincare disc model as the Euclidean map of the hyperbolic space. There the entire hyperbolic space is mapped into the interior of the sphere $\alpha^2 + \beta^2 + \zeta^2 = -\frac{4}{K_3}$. Visual straight lines are the circles intersecting the basic sphere orthogonally. The geodesics are symmetric to the $\alpha$ axis lying in the $\alpha - \beta$ plane and they are the circles normal to the basic circle $2/\sqrt{(-K_3)}$ as shown in Fig.4. Therefore considering the homeomorphism between the space proposed here and the space considered by Luneburg, we can consider either of Eq.28 or Eq.25 as the equation of horopters in the visual space and analysis of any one will hold good for the other under the Mobius transformation as explained above. To compare them with the experimental results we transform them in the physical Euclidean space using the Luneburg mapping function [2]. The local Cartesian coordinates $\alpha, \beta$ are related to the polar coordinates $\rho_1, \phi_1$ as,

$$\alpha = \rho_1 \cos(\phi_1) , \qquad (29a)$$

and $$\beta = \rho_1 \sin(\phi_1) . \qquad (29b)$$

It should be noted here that the local polar coordinates $\rho_1, \phi_1$ in the Poincare disc model possess the identical relations, as exists between the coordinates $\rho, \phi$ and $\{\mu, \sigma\}$ of the upper half plane model defined by the Fisher-Rao metric. Luneburg's mapping function between the variables in physical and visual space are defined as,

$$\rho_1 = 2\exp(-\tau\lambda) \qquad (30a)$$

$$\psi = \phi_1 \qquad (30b)$$

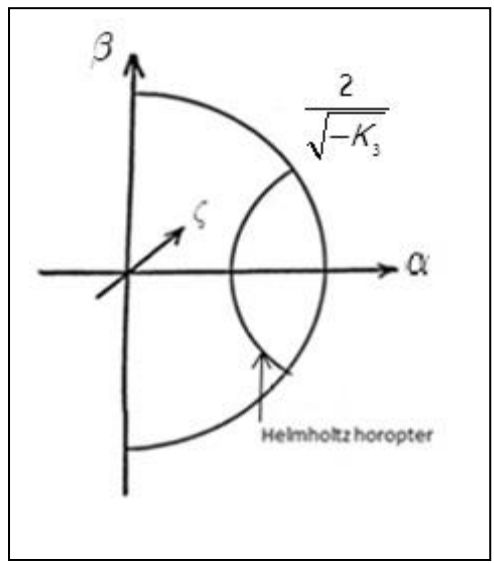

Fig.4. The Helmholtz horopter which appears as
Visual straight line in the hyperbolic visual space.

Here $\tau$ is an individual constant and varies among the subjects. It is related to the degree of depth perception of the observer. The mapping can be considered as purely ego-centric and rigid as described by Indow [26].Let us consider a point P in the Poincare disc model and the same is represented in the upper half plane as shown in Fig 1b.How far and in which direction the point is perceived with regard to the self in the visual space are solely determined by its bipolar coordinates in the physical space and regardless of the conditions of remaining points in the physical space. Therefore it's a context free mapping and holds only in the frameless condition. By the aid of Eq.29 and Eq.30 , Eq.28 can be expressed in terms of bipolar coordinates and personal constants as [2],

$$\frac{\cosh[\tau(\lambda + \kappa)]}{\cosh[\tau(\lambda_0 + \kappa)]} = \cos(\phi_1) \,, \tag{31}$$

where $\kappa = \dfrac{\ln\left(\dfrac{1}{-K_3}\right)}{2\tau}$ and $\lambda_0$ is the bipolar parallax on the x-axis. If the straight line horopter from the observer intersects the x-axis (Fig.3) at a point $L_0$ then it can be expressed as a function of the personal constant $\tau$ obeying the transcendental equation [2],

$$\tan\frac{\lambda_0}{2} \tanh[\tau(\lambda_0 + \kappa)] - \frac{1}{4\tau} = 0, \tag{32}$$

$$\text{and} \quad L_0 = \frac{\nu}{\lambda_0} \,. \tag{33}$$

Where $\nu$ is the interpupillary distance. Solving the transcendental Eq.32 one can compute the value of $\lambda_0$ and with the aid of Eq.33 the distance of the straight line horopter $L_0$ on the x-axis can be computed. Finally the coordinates in the physical space can be computed using the equations given below [ 26],

$$x = \frac{\nu}{2} \frac{\cos(2\phi_1) + \cos(\tau)}{\sin(\tau)} \,, \tag{34.a}$$

$$\text{and} \quad y = \frac{\nu}{2} \frac{\sin(2\phi_1)}{\sin(\tau)} \,. \tag{34.b}$$

The personal constant $\tau$, the interpupillary distance $\nu$ and the distance $L_0$ are measured experimentally [31]. All the experimental data along with the theoretical values computed using Eq.32 and Eq.33 are presented in Table 1. We further simulate the horopters below and above the straight line horopters using Eqn.34.a and 34.b. for three subjects. The personal constants are taken from the experimental data published in [31] and mentioned as A.J, T.K and S.V.S. The results are presented in Fig.5.

## 3B. Simulation of parallel and distance Alleys.

Distinction in visual observation between apparently parallel straight lines and curves of apparent equidistance is the most direct evidence advocating non-Euclidean characteristics of visual space. Blumenfeld [5], [6] first reported it through an experiment in a frame less scene, arranged using a dark room and a considerably large horizontal platform with the subject's eye fixed slightly above the plane. In Euclidean geometry equidistant and parallel lines are the same geometrical entities while in hyperbolic and elliptic geometry they are attributed with different forms. The physiological interpretation of equidistance can be stated as the length of the geodesics between pairs of stimuli are same. On the other hand the physiological interpretation of parallel lines in a space is not at all obvious or mathematically well posed. As explained by Luneburg [2] that parallelism can have different mathematical concepts and can be defended with equal good arguments. Unlike Euclidean geometry, hyperbolic or elliptic geometry does not offer absolute meaning to the statement that two line elements, not attached to the same points are parallel. Following Levi-Civita's concept one can use parallel transfer of a given line element along a given curve $\chi$ to another point Q parallel with itself. However the differential character of the space in non-Euclidean geometry the result of the transfer depends strongly on the curve $\chi$. If the nature of the curve changes the result of transfer will be different. At this juncture we hypothesize that the visual task of arranging stimuli against the instruction to make them equidistant is different from the task of arranging them as parallel lines. In the first case the observer's attention remains focused towards the interval between each pair of stimuli. While in case of parallel alleys the observer pays attention to watch the direction of lines consisting of rows of stimuli. It is found that the simulation of parallel alleys is obtained with larger values of $\sigma_1$ and $\sigma_2$ in comparison to the distance alleys. In both cases they are gradually increasing as the pair of stimuli come closer to the eye of the observer Fig.6.To compute the geodesic distance between each pair of stimuli Eq.11 is used and the widths $\sigma_1$ and $\sigma_2$ of the tuning curves are selected using the same method described in Sec.2A.

Table.1

| Observer | $\tau$ | Interpupillary distance $\nu$ (cm) | $L_0$, observed in horopter experiment (cm) | Theoretical value of $L_0$ obtained from Eqn.39 and 40 (cm). |
|---|---|---|---|---|
| A.J | 10.68 | 6.48 | 108 | 89.70 |
| T.K | 11.69 | 6.70 | 106 | 101.51 |
| M.K | 8.67 | 6.40 | 108 | 71.93 |
| M.V.S | 7.50 | 6.25 | 121 | 60.78 |
| B.A | 11.87 | 6.40 | 137 | 98.46 |
| S.V.S | 3.32 | 6.60 | 44 | 28.46 |
| H.S | 9.70 | 6.30 | 78 | 79.21 |
| K.G | 8.58 | 6.90 | 82 | 76.74 |
| W.K | 16.90 | 5.80 | 145 | 127.03 |

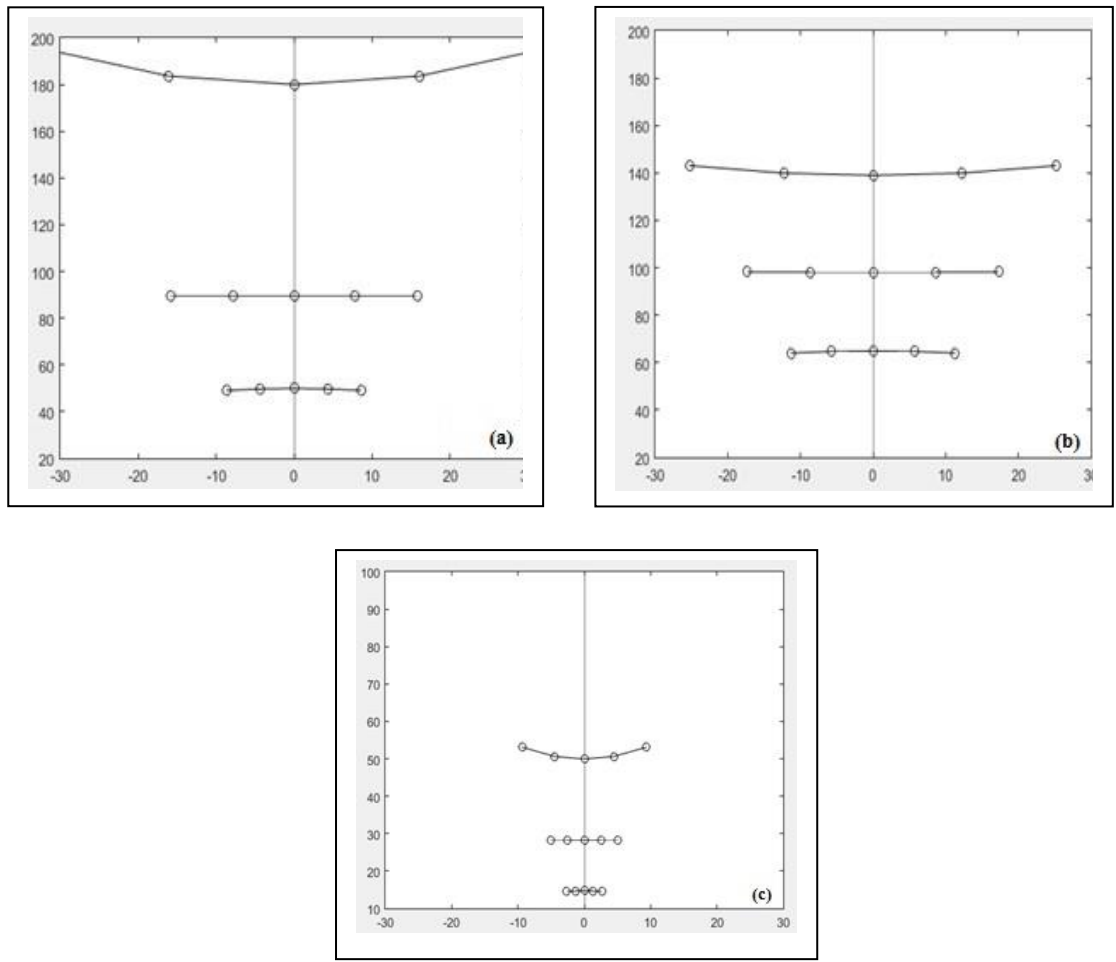

Fig.5 Simulated frontal plane horopters. Experimentally measured values of the personal constants like $\tau$, interpupillary distance $\nu$ are collected from [31]. (a) Represents the horopters for the observer, mentioned as A.J in [31] with $\tau=10.68$ and $\nu=6.48$ cm. The computed value of $L_0$ = 89.70. (b) Horopters for observer T.K with $\tau=11.69$, $\nu=6.70$ cm. and $L_0$ = 101.51.(c) Observer S.V.S., $\tau=3.32$, $\nu=6.60$ cm. and $L_0$ = 28.46. In each case the scalar curvature K of the visual space is taken as -.

The experimental data used in the simulation are collected from [31].

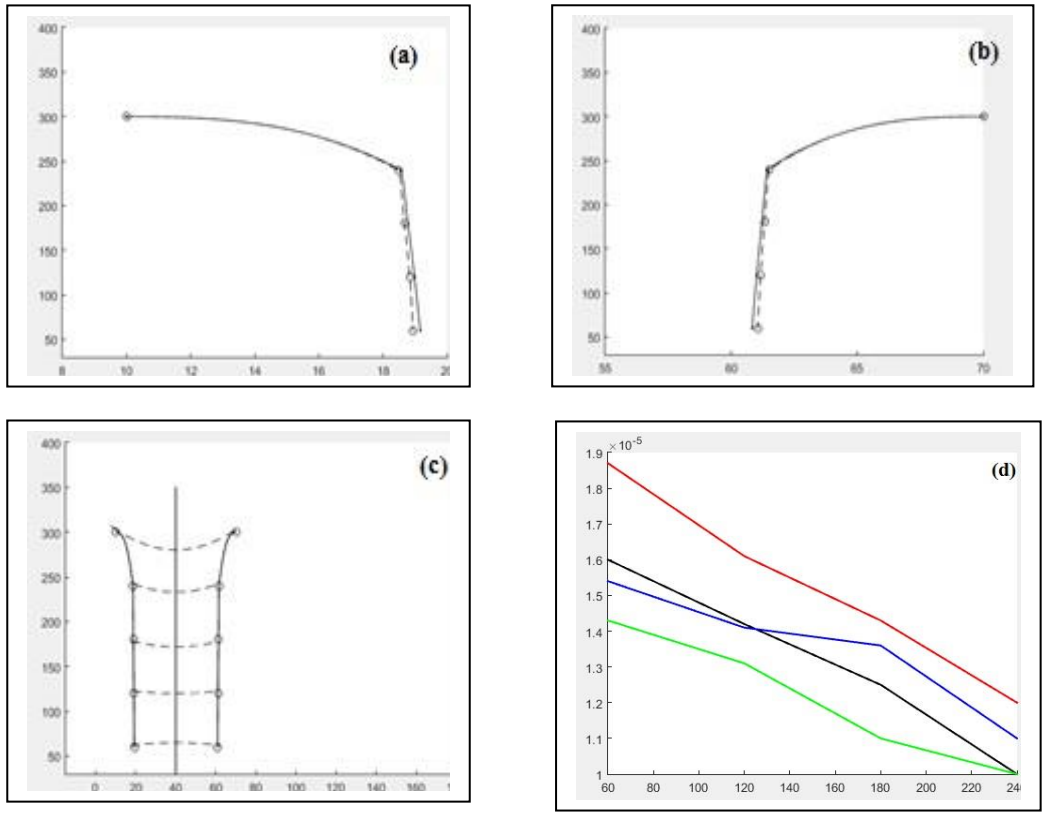

Fig.6. Computer simulated parallel, distance alleys and the horopters.(a) Parallel alley ( solid line) and the distance alley ( dotted line ) at the left side of the midline.(b) Parallel and distance alleys in the right part of the midline.(c) Parallel alleys intersected by frontal horopters ( dotted line).(d) Variation of $\sigma_1$ and $\sigma_2$ for parallel and distance alleys. Variation of $\sigma_1$ for parallel and distance alley are represented by black and green line. The red and blue line represents successively the variation of $\sigma_2$ for parallel and distance alley.

## 4. Conclusion

The present paper is an attempt to develop a model of the hyperbolic visual space formed at the visual cortex through a series of complex neural functions. The model is developed considering the physical existence of the visual space as a neural manifold. The response of each neuron in the manifold is represented by a response function ( tuning curve) and represented by Gaussian function. The choice of a probability space of Gaussian random variables to represent the neural manifold is considered to be the most appropriate. The Fisher-Rao metric is taken as a natural choice for the psychometric distance metric. Various geometric and topological properties of the space is described. And it is shown that the neural manifold modelled with Fisher-Rao information geometric space is a Riemannian hyperbolic space of constant negative curvature and it yields geodesics into it. Computer simulation of different visual phenomena associated with the hyperbolic visual space is presented. The simulation results are compared with published experimental data.

,